\documentclass[conference]{IEEEtran}

\usepackage{algorithm}
\usepackage{algpseudocode}
\usepackage{amsmath}
\usepackage{amsfonts}
\usepackage{xcolor}
\usepackage{booktabs}
\usepackage{multirow}
\usepackage{threeparttable}
\usepackage{graphicx}
\usepackage{url}
\usepackage{tablefootnote}
\usepackage{makecell}
\usepackage{pifont}

\usepackage{todonotes}

\newcommand{\cmark}{\ding{51}} % ✓
\newcommand{\xmark}{\ding{55}} % ✗

\begin{document}
\title{AIE4ML: An End-to-End Framework for Compiling Neural Networks for the Next Generation of AMD AI Engines}

\author{
  \IEEEauthorblockN{{
    Dimitrios Danopoulos\textsuperscript{1},
    Enrico Lupi\textsuperscript{1},
    Chang Sun\textsuperscript{1}
  }}

  \IEEEauthorblockN{{
    Sebastian Dittmeier\textsuperscript{3},
    Michael Kagan\textsuperscript{4},
    Vladimir Loncar\textsuperscript{2},
    Maurizio Pierini\textsuperscript{1}
  }}

  \IEEEauthorblockA{
    {\textsuperscript{1}European Organization for Nuclear Research (CERN), Geneva, Switzerland}\\
    {\{dimitrios.danopoulos,enrico.lupi,chang.sun, maurizio.pierini\}@cern.ch}
  }

  \IEEEauthorblockA{
    {\textsuperscript{2}Institute of Physics Belgrade, Serbia}\\
    {vloncar@ipb.ac.rs}
  }

  \IEEEauthorblockA{
    {\textsuperscript{3}Physikalisches Institut, Heidelberg University, Germany}\\
    {dittmeier@physi.uni-heidelberg.de}
  }

  \IEEEauthorblockA{
    {\textsuperscript{4}SLAC National Accelerator Laboratory, Menlo Park, CA, USA}\\
    {makagan@slac.stanford.edu}
  }
}

% make the title area
\maketitle

\begin{abstract}
  Efficient AI inference on AMD's Versal AI Engine (AIE) is challenging due to tightly coupled VLIW execution, explicit datapaths, and local memory management. Prior work focused on first-generation AIE kernel optimizations, without tackling full neural network execution across the 2D array. In this work, we present \texttt{AIE4ML}, the first comprehensive framework for converting AI models automatically into optimized firmware targeting the AIE-ML generation devices, also with forward compatibility for the newer AIE-MLv2 architecture. At the single-kernel level, we attain performance close to the architectural peak. At the graph and system levels, we provide a structured parallelization method that can scale across the 2D AIE-ML fabric and exploit its dedicated memory tiles to stay entirely on-chip throughout the model execution. As a demonstration, we designed a generalized and highly efficient linear-layer implementation with intrinsic support for fused bias addition and ReLU activation. 
  Also, as our framework necessitates the generation of multi-layer implementations, our approach systematically derives deterministic, compact, and topology-optimized placements tailored to the physical 2D grid of the device through a novel graph placement and search algorithm. Finally, the framework seamlessly accepts quantized models imported from high-level tools such as \texttt{hls4ml} or \texttt{PyTorch} while preserving bit-exactness. In layer scaling benchmarks, we achieve up to 98.6\% efficiency relative to the single-kernel baseline, utilizing 296 of 304 AIE tiles (97.4\%) of the device with entirely on-chip data movement. With evaluations across real-world model topologies, we demonstrate that \texttt{AIE4ML} delivers GPU-class throughput under microsecond latency constraints, making it a practical companion for ultra-low-latency environments such as trigger systems in particle physics experiments.
\end{abstract}

\section{Introduction}
%The rapid progress of deep learning has fueled an ever-increasing demand for specialized AI accelerators. 
At the algorithmic level, modern architectures focus on the General Matrix-Matrix Multiplication (GEMM), which dominates training and large-batch inference in models such as large language models (LLM) \cite{NEURIPS2020_1457c0d6} and Vision Transformers (ViT) \cite{Dosovitskiy2020AnII}. In contrast, many latency-sensitive inference workloads operate in a fundamentally different regime: during autoregressive LLM inference, for example, projection and feedforward layers are executed on a single token at a time and therefore reduce to General Matrix-Vector Multiplication (GEMV) or very low-batch GEMM; similar structures arise in State Space Models such as Mamba~\cite{Gu2023MambaLS} and in the linear layers of MLP-Mixers~\cite{10.5555/3540261.3542118}. These GEMV-like operators receive far less optimization attention than high-throughput GEMM, despite being the dominant compute path in real-time and token-by-token inference. Mainstream AI accelerators, such as GPUs and TPUs, are instruction-driven processors that excel at throughput-oriented workloads, which in turn are often not suitable for ultra-low-latency environments. For instance, the trigger systems of experiments at CERN require deterministic on-chip inference with microsecond-level response times \cite{hls4ml_paper1}. In these types of settings, FPGAs have demonstrated the value for highly or even fully unrolled matrix-vector operations in graph neural networks \cite{ExaTrkX:2020nyf} and compact MLPs \cite{hls4ml_paper2, que2025jedilinear}. Also, in such settings, low-precision arithmetic has become a key enabler for achieving high performance and energy efficiency in neural-network inference~\cite{low_quant_paper1, low_quant_paper2, 10980492, 9913212}. Yet, since FPGA on-chip resources are scarce and saturate quickly when parallelism is increased, scaling these designs to larger and deeper models remains challenging, leaving a significant gap between current solutions and the demands of next-generation of real-time inference.

The AMD Versal architecture could be a promising candidate to fill this gap. It integrates programmable logic (PL), a processing system (PS), and a two-dimensional array of AI Engines (AIEs), very-long intruction words (VLIW) vector processors with local memory and flexible interconnects. The second-generation AIE-ML architecture further improves the architecture with larger local memories, higher compute capacity, and additional memory tiles that are shared across all AIEs. Prior works, such as MaxEVA~\cite{maxeva} and AMA~\cite{ama}, primarily focus on highly optimized GEMM operators on AIE devices, while AutoMM~\cite{automm} and CHARM~\cite{10.1145/3686163} explore composing multiple matrix multiplication kernels on AIE devices as part of heterogeneous execution pipelines. However, these approaches do not easily extend to fully automated, end-to-end execution of complete neural networks, do not target the newer AIE-ML devices and and do not emphasize a fully on-chip dataflow across layers.

In this work, we present \texttt{AIE4ML}, the first end-to-end framework that seamlessly compiles complete neural networks from high-level frameworks such as PyTorch, TensorFlow, into optimized firmware for the new generation of AMD AIE-ML devices. The main contributions of this paper are:

\begin{itemize}
  \item First framework to fully exploit the AIE-ML architecture and its memory tiles for end-to-end, on-chip neural network execution.
  \item Highly efficient and generic linear-layer implementations with intrinsic support for fused bias and activation, achieving performance close to the architectural peak.
  \item A novel search algorithm that minimizes interconnect overhead between graphs for deterministic and optimized graph placement throughout the 2D AIE-ML array.
  \item Direct support for quantized and mixed-precision models imported from frameworks such as PyTorch and TensorFlow, preserving bit-exactness across the toolflow.
  \item Thorough evaluation across a variety of layer topologies and hardware architectures, showing GPU-class throughput under tight latency constraints.
\end{itemize}

\vspace{0.4cm}

\section{Related Work}

Almost all prior accelerator frameworks focused on targeting AMD's first-generation AIE architecture, mainly for single-kernel GEMM acceleration. As such, they did not address the system-level challenges of mapping multiple neural-network layers, handling inter-layer communication, or managing on-chip memory resources.

MaxEVA~\cite{maxeva} and AutoMM~\cite{automm} implement high-performance GEMM engines on first-generation AIE, while AMA~\cite{ama} provides an analytical model for predicting GEMM performance and identifying bottlenecks in AIE tile execution. While AutoMM supports composing multiple GEMM kernels, these works primarily focus on single GEMM operators, without mechanisms for chaining multiple layers, maintaining on-chip inter-layer dataflow, or building full neural-network graphs. With \texttt{AIE4ML} and through the use of memory tiles, the model design stays entirely on the AIE chip, avoiding the PL I/O overhead that other AIE solutions often assume away. CHARM~\cite{10.1145/3686163} partially addressed the challenge of models with many layers of varying dimensions by composing heterogeneous MatMul accelerators on AIE and PL. Nevertheless, CHARM still relies heavily on PL-side buffering, aggregation and control which limits the fully on-chip, multi-layer dataflow. ARIES~\cite{10.1145/3706628.3708870} introduced an MLIR-based programming model for AIE and PL. While useful, ARIES still requires users to explicitly map kernels onto AIE cores, select tile sizes, and specify task-level parallelism and placement. Its evaluation, similar to MaxEVA and AutoMM, focused primarily on GEMM kernels rather than complete multi-layer neural-network graphs. On the tooling side, AMD's IRON \cite{iron} provides a “close-to-metal” programming API for AIE/NPU devices with support for placement and vector-level programming. However, IRON is a low-level expert-oriented toolkit: scaling workloads across the 2D array, finding correct and optimized placement, and managing buffer tiling are left to the designer. Also, its support for mixed-precision neural network inference is not well documented. In contrast, \texttt{AIE4ML} provides explicit support for the integer precisions commonly used on AIE-ML. Finally, GAMA~\cite{gamma} is the latest framework targeting AIE-ML, achieving high GEMM throughput. Like MaxEVA and AutoMM, GAMA focuses on a single high-performance GEMM operator, not on constructing multi-layer inference pipelines. As a result, it does not address bias or activation inclusion, inter-layer communication, or most notably, on-chip data rearrangement via Memory tiles, which is one of the defining features of the AIE-ML architecture.

In summary, none of the existing frameworks provide an end-to-end flow that (i) targets AIE-ML, (ii) ingests neural networks directly from high level tools (i.e., PyTorch), (iii) ensures bit-exact quantized inference, (iv) automatically determines connections, parallelization, and placement across the 2D AIE-ML array, or (v) executes all layers entirely on-chip through Memory Tiles without PL involvement. \texttt{AIE4ML} is the first framework to offer this fully automated, graph-level, on-chip compilation pipeline for AIE-ML devices.

\section{Hardware design}

In \texttt{AIE4ML}, networks are constructed by chaining layer-level graphs through memory tiles. Scaling within a layer is achieved by replicating the corresponding kernels across tiles, while scaling across layers is achieved by distributing subgraphs over the 2D AIE array. Data movement between layers is handled by memory tiles, which serve as shared buffers and connectors that store intermediate activations and redistribute them to successive layers. Double buffering, also known as ping-pong buffering, is employed in the memory tiles as well as in the input/output buffers of the AIE tiles. This allows one buffer to be filled or drained while the other is being used by the kernel, overlapping communication and computation to maintain throughput and avoid stalls. While, this design is explicitly optimized for the AIE-ML architecture, its modular structure makes it easy to adapt to the more recent AIE-MLv2 generation.

Our framework follows a neural-network-oriented execution model in which the weights and biases are loaded once from the processing system (PS) through run-time parameter (RTP) ports and stored directly in local AIE memories. Unlike prior AIE-based accelerators that always require streaming all matrix operands from the PL \cite{gamma, maxeva}, \texttt{AIE4ML} keeps the weight parameters resident on-chip throughout inference. This effectively removes the PL-side bandwidth bottlenecks and provides flexibility by enabling model updates such as new coefficients. The remainder of this section details the kernel, memory-tile, and scaling mechanisms.

\begin{figure*}[t] % use [t] for top placement
  \centering
  \includegraphics[width=1.05\textwidth]{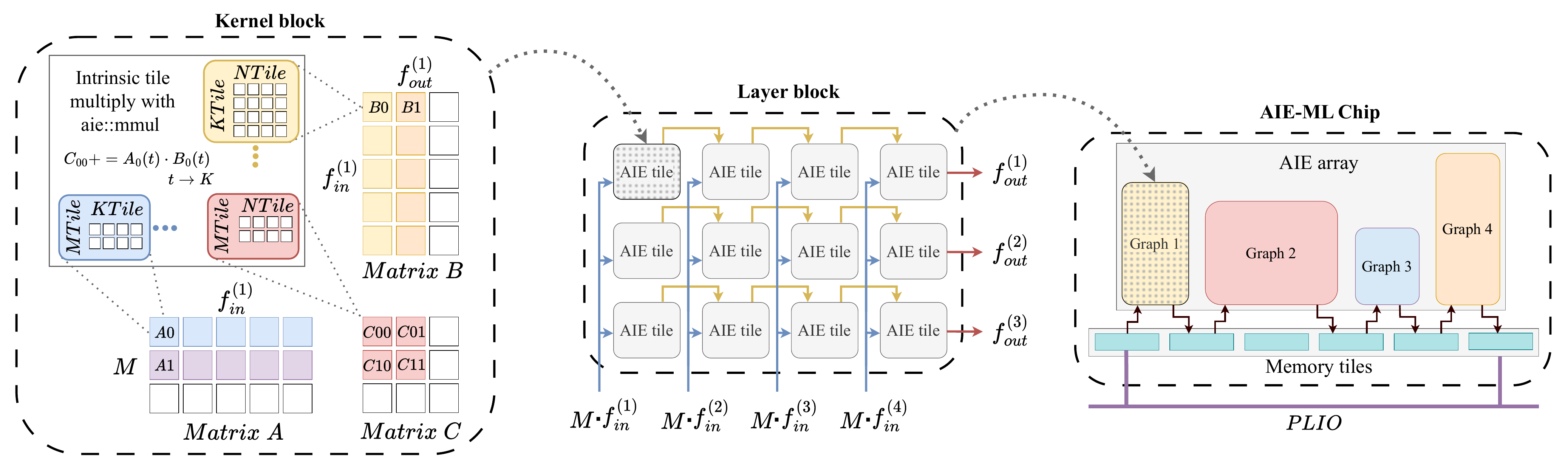}
  \caption{
    Overall hardware design of \texttt{AIE4ML}.
    \textbf{Left:} Blocked layer kernel using the \texttt{aie::mmul} API and the 2\(\times\)2 accumulator scheme, shown here with an illustrative \(\langle 2,4,4\rangle\) tile size.
    \textbf{Middle:} Layer-level scaling using cascade rows and input broadcasting to distribute inputs and combine partial sums across the AI engine array.
    \textbf{Right:} Cross-layer pipelining through memory tiles, which provide zero-padding, re-tiling, and activation distribution to enable fully on-chip multi-layer execution.
  }
  \label{fig:design}
\end{figure*}

\subsection{Kernel and memory design}
At the core of computation is the AI Engine kernel, implemented as a C++ program that leverages intrinsic operations supported by the compiler and executes on a single AIE tile. While there are multiple ways to design a linear layer for the AIE, we adopt the use of the \texttt{aie::mmul} API because it offers robust performance, general applicability, and simplifies maintaining data alignment requirements. Unlike dot-product implementations that require very wide consecutive vectors of input elements (e.g., 32 int8 values for a single multiply-accumulate), the \texttt{aie::mmul} formulation performs block-level matrix multiplication, distributing the dot products across multiple dimensions. Conceptually, we see the classical GEMV operation as a matrix-matrix multiplication with a batch size of one. In this way, increasing the batch size corresponds to adding more rows to the input matrix, which allows the AIE kernels to fully exploit the accumulator lanes of their MAC unit and push the performance even more, especially when GEMV operation has short vector lengths.

The \texttt{aie::mmul} class template is parameterized by ($M \times K \times N$), the input and output data types, and optionally the accumulation precision. Once instantiated, it interprets input matrices as contiguous vectors with the given tile sizes and provides a multiply-accumulate interface that yields results convertible to accumulators or vectors. In our linear layer kernel, we exploit this by computing two input tiles (rows of $A$) against two weight tiles (columns of $W$) per iteration. This produces four accumulators $C_{00},\,C_{01},\,C_{10},\,C_{11}$ in parallel, effectively unrolling across both the batch and output-tile dimensions (as shown on the left of Figure~\ref{fig:design}). This blocked schedule increases arithmetic intensity by reusing each loaded tile of $A$ and $W$ across multiple accumulators, hence amortizing the apparent load imbalance frequently seen in single-tile schedules. With multiple active accumulators, kernel prologue/epilogue bubbles are overlapped for steady throughput. Finally, we chose to use the \texttt{io\_buffer} interface, which provides direct access to the local memory banks with wide vector loads and stores. This interface guarantees that computation is continuously fed at the rate required to keep the MAC units fully utilized. While the $2{\times}2$ scheme is usually efficient on AIE-ML devices, using more blocks can improve accumulator usage on AIE-MLv2 devices.

\textbf{Optimized VLIW Execution:}
The linear layer kernel is structured as a blocked matrix multiplication that maps naturally to the AIE-ML's
7-way VLIW execution model. Each iteration attempts to schedule one vector multiply-accumulate (\texttt{VMAC}),
two vector loads (\texttt{VLDA}, \texttt{VLDB} from each load unit), one vector store (\texttt{VST}), one scalar address update, and move
operations, thereby realizing near II=1 throughput. By unrolling over both the batch and output
dimensions, the kernel computes four accumulator blocks in parallel
($C_{00},C_{01},C_{10},C_{11}$), which hides memory latency while stores overlap with subsequent multiplies.
%Stores overlap with subsequent multiplies, while scalar pointer arithmetic occupies the  AGU slot, so that the VLIW processor remains busy all time. 
Best practices at the microcode level are applied explicitly:
\begin{itemize}
  \item Input/output buffers are 32-byte aligned, mapped to distinct banks so that parallel loads and stores are served by different banks.
  \item Weights/biases are RTP-loaded and resident on-chip for flexible updates; input/output buffers are double-buffered to overlap computation with data movement.
  \item Quantization is fused into the store: \texttt{VST.SRS} applies shift (scaling), rounding, and saturation in one step.
  \item $(M,K,N)$ tile dimensions are chosen to match native AIE-ML vector widths, avoiding extra instructions for data manipulation.
  \item Bias values, when enabled, are loaded into the accumulators during initialization (prologue),
        while optional ReLU activation is applied in the epilogue prior to the final vector store.
\end{itemize}
The resulting schedule follows the expected prologue (initial \texttt{VMUL} operations after the initial loads of data), steady-state loop (two loads + one MAC + scalar update per cycle), and epilogue (scaling, optional activation, and stores), which together constituting full pipeline utilization (see Algorithm~\ref{alg:aie_dense_algo}).

\begin{algorithm}[t]
  \caption{Pseudocode of AIE Linear kernel}
  \label{alg:aie_dense_algo}
  \begin{algorithmic}[1]
    \footnotesize
    \Require Input buffer $A$, weight buffer $W$, optional bias $b$, tile shape $(M,K,N)$, shift $s$, flags: \textsc{UseBias}, \textsc{UseReLU}
    \Ensure Output buffer $C$
    \For{each $M$-row tile of $A$}
    \For{each $N$-column tile of $W$}
    \State \textbf{ACC\_INIT}: allocate accumulators $C_{00},C_{01},C_{10},C_{11}$
    \If{\textsc{UseBias}}
    \State \textbf{BIAS\_LOAD}: fetch bias tile, replicate across accumulators
    \EndIf
    \For{each $K$-slice along reduction}
    \State \textbf{VLDA}: load $M\times K$ tile from $A$ into vector register
    \State \textbf{VLDB}: load $K\times N$ tile from $W$ into vector register
    \State \textbf{VMAC}: update accumulators with multiply-accumulate
    \EndFor
    \State \textbf{SRS}: apply shift/round/saturate
    \If{\textsc{UseReLU}} \State \textbf{ACTIVATION}: apply ReLU clamp \EndIf
    \State \textbf{VST}: store output tile to $C$
    \EndFor
    \EndFor
  \end{algorithmic}
\end{algorithm}

\textbf{Performance Bound Analysis:} To contextualize the measured performance, we first establish the maximum theoretical throughput of a single AIE-ML tile, under $f = 1.25$\,GHz. Each AIE can issue one \texttt{VMAC} / cycle, together with two 256-bit loads and one 256-bit store. Therefore:
\begin{equation}
  \text{Peak}_{\text{compute}} = W(p_A,p_B)\, \times f \quad [\text{MAC/s per AIE tile}],
\end{equation}
where $W(p_A,p_B)$ denotes the number of parallel MACs per cycle supported for the precision pair (e.g., $W(\text{8b},\text{8b})=256$).

Another limitation comes from the memory bandwidth: with two 256-bit load ports, at most 64\,B/cycle can be loaded, corresponding to only $\approx 32$\,MAC/cycle (int8) in the absence of reuse. For GEMV (batch size = 1), reuse is limited and this memory ceiling dominates. Blocked formulations such as \texttt{aie::mmul} increase this bound by reusing activations across multiple batch elements (and, to a lesser extent, across output tiles), thereby amortizing memory traffic and moving steady-state performance closer to the architectural bound.

The achievable peak performance also depends on the chosen \texttt{aie::mmul} tiling, since per-tile efficiency is limited by the slowest stage among \texttt{VLDA}, \texttt{VLDB}, or \texttt{VMAC}. Our tool supports a wide range of tiling configurations, but for this study we selected a representative subset of native tilings that map directly to hardware intrinsics. Non-native tilings often require emulation through multiple intrinsic calls and extra data manipulation, which lowers efficiency. Even some native tilings appear suboptimal in isolation when loads or compute require more than one cycle per tile. However, with our 2$\times$2 blocked kernel each loaded tile is reused across four accumulators, amortizing this overhead and sustaining performance close to the architectural ceiling. Table~\ref{tab:theoretical-peaks} summarizes the selected \texttt{aie::mmul} tilings and their theoretical ceilings on a single AI engine (consistent with AMD's published performance table~\cite{amd_aie_perf}), which serve as the baseline for efficiency results in the evaluation section.

\begin{table}[h]
  \centering
  \caption{Single AIE-ML tile ceilings for selected tilings and integer datatypes at 1.25\,GHz.}
  \label{tab:theoretical-peaks}
  \footnotesize
  \begin{tabular}{cccccc}
    \toprule
    $\langle M,K,N\rangle$ & \textbf{Datatype} & \textbf{Native} & \textbf{MAC/cyc} & \textbf{GMAC/s} & \textbf{GOP/s} \\
    \midrule
    $\langle4,8,8\rangle$  & i8$\times$i8      & Yes             & 256              & 320             & 640            \\
    $\langle4,4,8\rangle$  & i16$\times$i8     & Yes             & 128              & 160             & 320            \\
    $\langle4,4,4\rangle$  & i16$\times$i16    & Yes             & 64               & 80              & 160            \\
    \bottomrule
  \end{tabular}
\end{table}

\subsection{Scaling to Layer-Level Parallelism}
While a single kernel can compute a block of outputs, large layers with many input and output features quickly exceed the compute and memory limits of a single AIE tile. To address this, workloads are distributed horizontally and vertically across the 2D AIE-ML array. Dot-product accumulations are propagated west-to-east through the dedicated 512-bit cascade ports, so that partial sums are reduced across multiple tiles; at the same time, cascade rows are replicated north-south, so that each row produces a distinct slice of the output features in parallel. To avoid redundant data movement and additional wiring on the chip, input vectors are injected once per cascade column and broadcast upward from the memory tile to all compute tiles in that column (as illustrated in the middle of Figure~\ref{fig:design}).

In practice, we express scaling in terms of input and output slices. Each cascade row spans $\texttt{CAS\_LEN}$ tiles horizontally, with each tile consuming $f_{\text{in}}^{\text{slice}}$ input features. The total input dimension covered is therefore:
\[
  f_{\text{in}} = \texttt{CAS\_LEN} \cdot f_{\text{in}}^{\text{slice}}
\]
Similarly, the array is partitioned vertically into $\texttt{CAS\_NUM}$ cascade rows, each producing $f_{\text{out}}^{\text{slice}}$ output features, so that the total output dimension is
\[
  f_{\text{out}} = \texttt{CAS\_NUM} \cdot f_{\text{out}}^{\text{slice}}
\]

\textbf{Data Partitioning through Memory tiles:} Each compute tile executes a linear layer kernel on local dimensions $f_{\text{in}}^{\text{slice}}$ and $f_{\text{out}}^{\text{slice}}$, without visibility of the global layer shape. The sequencing of these blocks is orchestrated by the AIE-ML memory (MEM) tiles, whose direct memory access (DMA) engines deliver streams of tiles according to programmable tiling parameters. These parameters specify (i) the \emph{buffer dimension}, i.e., the full logical extent of the stored buffer, (ii) the \emph{tiling dimension}, i.e., the inner block dimension of the data transfer in the buffer, and (iii) the \emph{tile traversal}, defined by stride and wrap. The stride determines the distance in terms of buffer element data type between consecutive inter-tile traversal in this dimension, while the wrap specifies how many tiles to access on each dimension ~\cite{am020}. In this way, we leveraged MEM tiles to reorder and partition activations into the precise sequence of $M{\times}K$ and $K{\times}N$ blocks expected by $\texttt{aie::mmul}\langle M,K,N\rangle$, thus allowing kernels to remain agnostic of the global tensor layout and keeping execution fully on-chip without the need of the PL to aggregate or re-arrange data.

\subsection{Scaling Across Layers and Graph Interconnect}
Typically, neural networks require mapping multiple layers across the array, which we carried out by chaining AIE graphs through MEM tiles. Each layer is represented as a graph of compute kernels and their local buffers, while MEM tiles provide the glue between these graphs. Specifically, as in local input/output buffers, we also implement ping-pong buffering in the MEM tiles so that one buffer is refilled while the other drains, allowing communication to overlap with computation. Since MEM tiles expose independent write and read tilers, they naturally supported re-tiling of activations between the layers. For instance, a $layer_{i}$ may write results in $\{M_i,\,N_i\}$ tiles, while the subsequent $layer_{i+1}$ may read them in $\{M_{i+1},\,K_{i+1}\}$ tiles. This allowed us to handle mixed-precision execution across layers, where consecutive layers may operate on different data types and therefore require different tiling/layout configurations. To handle arbitrary layer dimensions, we also exploit the built-in zero padding of MEM tiles, where the DMA injects zeros when accessing data outside the defined buffer boundaries. This flexibility allows arbitrary layer shapes and tiling schemes to be connected without modifying kernel code. Finally, we use the broadcasting mechanism through the MEM tiles to stream activations vertically to all compute tiles of the same column, thereby reducing routing overhead. Overall, this design enables layer-to-layer pipelining at high throughput and low latency.

\section{Design of \texttt{AIE4ML} framework}
The \texttt{AIE4ML} framework provides an end-to-end toolchain that automatically produces optimized AIE firmware directly from a high-level neural network description. Starting from quantized models, we reuse the frontend parser provided by \texttt{hls4ml} \cite{hls4ml_arxiv}, a free and open-source software commonly used in low-latency ML applications. \texttt{AIE4ML} then generates a ready-to-build AIE project that can be compiled and simulated using the AMD Vitis environment. We built this framework around three core principles: (1) an Intermediate Representation (IR) in \texttt{AIE4ML} designed to capture AIE-specific concepts; (2) a set of modular passes - each of which progressively populates the attributes of the IR nodes; and (3) a templated code generation system that instantiates optimized kernel and graph definitions for each layer.

\begin{figure}[t!]
  \centering
  \includegraphics[width=\columnwidth]{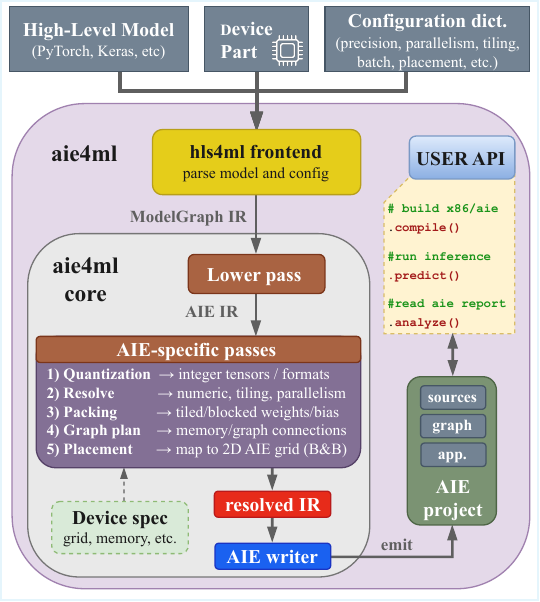}
  \caption{Overview of the \texttt{AIE4ML} compilation pipeline. A high-level network is parsed with \texttt{hls4ml} and then lowered into an AIE-specific intermediate representation (IR), processed by a sequence of passes that resolve quantization, tiling, packing, graph connectivity, placement, and finally emitted as an optimized AIE project ready for build or simulation.}

  \label{fig:aie4ml}
\end{figure}

\subsection{Framework operation}
The \texttt{AIE4ML} framework is based on a structured compilation pipeline, which progressively lowers a high-level neural network down to a fully resolved, device-aware AIE firmware. Figure~\ref{fig:aie4ml} provides a high-level overview of this flow.

\textbf{Intermediate Representation (IR):}
At the heart of this framework is a dedicated intermediate representation (IR) that captures AIE-specific attributes. During lowering, the \texttt{hls4ml} IR graph is transformed into this separate AIE-IR graph where each node represents an operation with embedded metadata on layer topology, tensor dimensions, quantization, and connectivity. This IR is then used as the central data structure for all subsequent passes, which populate its attributes with additional information regarding quantization, packing, placement, and code generation. Inferred attributes can be overridden by the user configuration directives; for example, bitwidths, cascade parameters, tiling shapes or placement coordinates, provided they are valid for the target device and design.

\textbf{Pass Pipeline:}
Model transformation is organized as a series of compiler passes, each consuming  enriching the IR:
\begin{enumerate}
  \item \emph{Lowering} creates the \texttt{AIE4ML} IR, applies simple fusions (e.g., Dense+ReLU), and initializes device context.
  \item \emph{Quantization} converts tensors into supported integer representations for target AIE device.
  \item \emph{Resolve} derives all deterministic AIE attributes, such as numeric types, parallelism, tiling, and placement, but honors any user-defined attributes that are valid.
  \item \emph{Packing} reorganizes quantized stationary tensors (weights and biases) into tiled and aligned layouts compatible with the formats expected by AIE intrinsics.
  \item \emph{Graph-planning} stage determines the explicit connections
        between compute graphs and memory tiles.
  \item \emph{Placement} maps layers onto the physical 2D AIE grid using the branch-and-bound search (described in  \ref{subsec:graph_placement}).
  \item \emph{Project Emission} instantiates pre-defined C++ layer templates using the resolved configuration of each node and renders the top-level graph, main application, and header files through Jinja. The result is a ready-to-build project targeting the AMD Vitis toolchain.

\end{enumerate}

\subsection{Toolflow and Simulation}

At the front end, users provide a quantized model (e.g., from \texttt{PyTorch} or \texttt{Keras}) through \texttt{hls4ml}, specifying the backend as \texttt{'AIE'}. The backend is registered via the \texttt{hls4ml} plugin system, so model conversion follows the same interface as existing FPGA flows. Additional directives related to precision, tiling, parallelization factors or graph placement can be supplied through the \texttt{hls4ml} configuration interface. Based on these configurations, the framework automatically makes decisions regarding numeric types, parallelization, and placement strategies for the target device. The flow supports both fast x86 simulation and hardware-level AIE simulation. The x86 mode allows fast functional validation, while the AIE mode performs a more accurate verification of the vector operations and memory accesses. Inference is executed seamlessly through the standard \texttt{predict()} interface of the generated AIE graph, with the user specifying the execution mode, (\texttt{x86} or \texttt{aie}). The AIE hw simulation further reports detailed profiling metrics including throughput, tile utilization, latency, and other hardware-level statistics. While the generated AIE project operates natively on quantized integer data types, users may optionally enable input quantization and output dequantization to perform inference directly with floating-point NumPy arrays. The resulting outputs are bit-exact with respect to the quantized \texttt{hls4ml} model.

\subsection{Graph placement on the 2D array} 
\label{subsec:graph_placement}

We visualize each layer (or graph) $G_i$ as a rectangular block of width equal to the cascade length and height equal to its cascade count. Let $(c_0, r_0)$ be the starting coordinates for $G_0$, then subsequent graphs are placed with the objective to minimize the weighted cost:

\begingroup
\setlength{\abovedisplayskip}{3pt}
\setlength{\belowdisplayskip}{3pt}
\begin{equation}
  J = \sum_{i} \Bigl( |c_\text{out}^i - c_\text{in}^{i+1}| \;+\;
  \lambda |r_\text{out}^i - r_\text{in}^{i+1}| \;+\; \mu \, r_\text{top}^i \Bigr),
  \label{eq:transition_cost}
\end{equation}
\endgroup

where $c_\text{in,out}^i$ and $r_\text{in,out}^i$ denote the input/output column and row of $G_i$ respectively, $r_\text{top}^i$ refers to the topmost row occupied by the block, and $(\lambda,\mu)$ are user-defined weights to control the placement heuristics. Our algorithm performs a branch-and-bound (B\&B) search that enumerates feasible, non-overlapping placements in bounds, incrementally accumulates $J$, and prunes partial assignments as soon as knowing that they cannot improve upon the current best. Constrained coordinates specified by the user are treated as hard constraints and the solver respects explicit overrides while still optimizing the placement of the remaining graphs. Algorithm output constitutes a legal tiling of the computational graph that aligns cascades horizontally, discourages large vertical hops, and biases the layout towards lower rows where buffering resources aggregate in the shared memory tiles. The algorithm’s runtime is negligible for networks that fit within the AIE array, typically requiring only a few seconds to generate near-optimal placements.

\begin{figure}[t!]
  \centering
  \includegraphics[width=0.8\columnwidth]{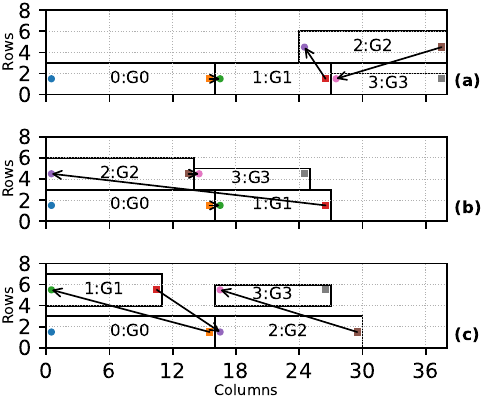}
  \caption{Automatic placement based on B\&B algorithm (a) compared with two greedy baselines (b,c) on a 38\,$\times$\,8 AIE array (start at $(0,0)$, $\lambda{=}1.0$, $\mu{=}0.05$). B\&B yields shorter inter-layer connections and lower-row bias.}
  \label{fig:aie-bnb-vs-greedy}
  %\vspace{-2mm} % (optional) tighten vertical spacing if needed
\end{figure}

Figure \ref{fig:aie-bnb-vs-greedy} shows example placements of the same graphs generated by our enumerative (B\&B) placement strategy  and two alternative greedy strategies. Greedy simple heuristics, such as always placing the next graph immediately to the right or directly above the previous one, would lead to legal but inefficient layouts, with either long horizontal traversals or unnecessary vertical hops. In contrast, by exploring feasible alternatives and evaluating them with the objective in Eq.~\eqref{eq:transition_cost}, B\&B discovers compact tilings that align cascades and remain on lower rows, which shortens inter-layer connections while respecting resource constraints. While manual placement could be done for toy cases, it rapidly becomes cumbersome and error-prone as graphs scale in depth and width. Automating this pass ensures consistent, near-optimal mappings across workloads and eliminates ad-hoc decisions that compound inefficiencies in complex designs.

\section{Evaluation}

The following performance results for \texttt{AIE4ML} were obtained using the cycle-accurate AIE simulator provided in the AMD Vitis~2025 toolchain. All AIE benchmarks target the AIE-ML generation of Versal devices
(VEK280 platform). \texttt{AIE4ML} is also compatible with the newer AIE-MLv2 architecture (has been functionally validated with the VEK385 platform), but we report results on AIE-ML due to its more stable and mature toolchain and performance model. We evaluate our framework across single-kernel microbenchmarks with mixed precision, multi-layer MLP-Mixer blocks, and arbitrarily stacked MLPs, and compare its performance against prior AIE-based frameworks, as well ass the state-of-the-art FPGA, GPU, and Apple NPU toolflows.

\subsection{Single kernel performance}

We begin by characterizing the performance of a single linear layer generated by
\texttt{AIE4ML}. The kernel is executed with a large batch (usually $B{=}128$) on a single compute tile for multiple
iterations to measure sustained throughput. Results are reported in GOPS (Giga Operations Per Second) and compared against the theoretical ceilings defined in Table~\ref{tab:theoretical-peaks}. For completeness, we also measure the end-to-end latency (including AIE I/O) of the base kernel at a micro-batch setting (i.e., $B{=}8$) and $4\times4$ cascade configuration. Although single-sample execution ($B=1$) is also supported, $B=8$ already saturates the achievable minimum latency by amortizing fixed per-inference pipeline and I/O overheads. Table~\ref{tab:gemv-throughput} summarizes the results across different workloads and datatypes.

\begin{table}[t!]
  \centering
  \caption{Single-kernel performance for different input precisions. Throughput (GOPS, efficiency in parentheses) and latency.}
  \label{tab:gemv-throughput}
  \footnotesize

  \begin{threeparttable}

  \begin{tabular}{lcccc}
    \toprule
    \textbf{Datatype} & \textbf{Workload} &
    \textbf{Base} & \textbf{+Bias+ReLU} & \textbf{Latency} \\
    \midrule
    i8 $\times$ i8$^{*}$ &
    $128{\times}128$ & $613$ (95.8\%) & $520$ (81.3\%) & $0.5\,\mu\text{s}$ \\
    \midrule
    i16 $\times$ i8$^{*}$ &
    $128{\times}128$ & $314$ (98.1\%) & $287$ (89.7\%) & $0.6\,\mu\text{s}$ \\
    \midrule
    i16 $\times$ i16$^{\dagger}$ &
    $64{\times}64$ & $154$ (96.3\%) & $114$ (71.3\%) & $0.5\,\mu\text{s}$ \\
    \bottomrule
  \end{tabular}

  \begin{tablenotes}[flushleft]
    \footnotesize
    \item[*] Uses 32-bit accumulator, 32-bit bias and stores outputs in 8-bit.
    \item[$\dagger$] Uses 64-bit accumulator, 32-bit bias and stores outputs in 16-bit.
    \item {Tilings:}\;
      i8$\times$i8: $\langle4,8,8\rangle$ \quad
      i16$\times$i8: $\langle4,4,8\rangle$ \quad
      i16$\times$i16: $\langle4,4,4\rangle$
  \end{tablenotes}

  \end{threeparttable}

\end{table}

\subsection{Model Performance}
\label{subsec:model-performance}

\begin{figure}[t!]
  \centering
  \vspace{0.2cm}
  \includegraphics[width=\linewidth]{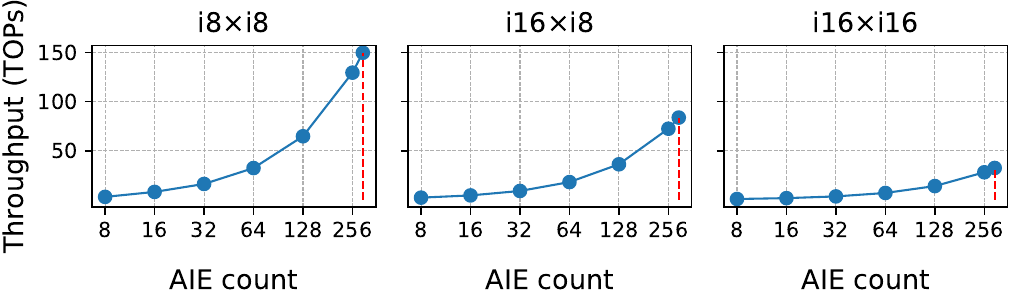}
  \caption{Scaling of a single linear layer (including bias and ReLU) across increasing AIE tiles. Red dashed lines mark maximum utilization at 296 of 304 tiles (97.4\% utilization).}
  \label{fig:weak-scaling}
\end{figure}

We further evaluate the scalability of \texttt{AIE4ML} across the full AIE-ML array. We take a single linear layer (including bias and ReLU) and scale it from a single AIE tile to the full array for each input datatype configuration in Table~\ref{tab:gemv-throughput}: i8$\times$i8, i16$\times$i8, and i16$\times$i16, distributed across varying cascade lengths and counts. The input size is increased proportionally with the number of AIE tiles. All data movement, including memory tile communication and data reordering, is handled entirely on-chip without PL involvement. Figure~\ref{fig:weak-scaling} reports absolute throughput as the number of AIE tiles increases. To evaluate scaling efficiency, we compare these values to the single-tile kernel from Table~\ref{tab:gemv-throughput}. Given sufficient input size, \texttt{AIE4ML} achieves near-ideal scaling across all three precisions, 97.3\%, 98.6\%, and 97.1\% for i8$\times$i8, i16$\times$i8, and i16$\times$i16, respectively. We highlight that \texttt{AIE4ML} can utilize up to 296 of 304 AIE tiles on the device (97.4\% spatial utilization), indicated by the red dashed lines in the figure. 
To contextualize these scaling results relative to the hardware peak, we additionally evaluate \texttt{AIE4ML} under a GEMM-only workload at full array utilization. In this configuration, \texttt{AIE4ML} sustains 160~TOPS, corresponding to 82.2\% of the theoretical INT8 peak of the AIE-ML device. Taken together, these results show that compute-bound configurations remain highly efficient. This study reflects scaling under realistic workloads, which often span many AIE tiles across the device, including memory tile overheads for data aggregation and reordering. This single-layer setup serves as a scalable building block that \texttt{AIE4ML} replicates across the array, enabling modular graph construction with consistent performance.
%Finally, while not shown in this figure, efficiency may degrade at smaller batch sizes as expected, due to limited parallelism and cascade backpressure. 

We also evaluate \texttt{AIE4ML} on a variety of layer and model configurations, including real-world models such as MLP-Mixers \cite{10.5555/3540261.3542118}, which are widely used in both general machine learning and real-time applications like fast jet tagging at CERN \cite{mlp_mixers_jet_tagging}. Each MLP-Mixer block consists of two linear sub-networks, \emph{token mixing} and \emph{channel mixing}, which apply learned transformations across different tensor dimensions. In our implementation every linear layer is immediately followed by a fused ReLU activation, both within Mixer MLPs and standalone MLP layers. Given an input tensor $X \in \mathbb{R}^{B \times T \times C}$, where $B$, $T$, and $C$ are the batch size, the number of tokens, and the number of channels, respectively, token mixing applies a linear map to each channel independently, while channel mixing applies it to each token. To implement these operations, in \texttt{AIE4ML} we can reshape the input as $[B \cdot C,\,T]$ for token mixing and $[B \cdot T,\,C]$ for channel mixing, yielding a GEMM-based execution. As shown in Table~\ref{tab:mixer-layers}, some MLP configurations experience slight performance degradation due to architectural constraints, such as 32-bit alignment requirements on tile or I/O boundaries. These restrictions limit how tensor dimensions can be sliced or cascaded, and may introduce padding overhead when shapes are not divisible by the hardware word size, reducing effective utilization. Also, when resources permit, the MLP block can be replicated across the AI Engine array to increase effective throughput. In general, \texttt{AIE4ML} sustains high throughput, especially in medium-sized models that are common in strict low-latency applications. Table~\ref{tab:mixer-layers} reports the sustained throughput and the steady-state per-sample output interval. We measure the time between consecutive full-batch outputs in steady state and divide by the batch size. All results are obtained with fully on-chip execution, including data reordering, tiling, and aggregation operations required during inference.

\setlength{\tabcolsep}{3pt}
\begin{table}[t!]
  \centering
  \caption{Performance of MLP-Mixer and standalone MLP blocks using \texttt{AIE4ML} (fully on-chip execution).}
  \label{tab:mixer-layers}
  \footnotesize
  \begin{tabular}{lccc}
    \toprule
    \textbf{Operation}  &
    \textbf{MOPs}      &
    \begin{tabular}{c}
      \textbf{Output Interval /} \\
      \textbf{Sample} ($\mu$s)
    \end{tabular} &
    \begin{tabular}{c}
      \textbf{Throughput} \\
      (TOPS)
    \end{tabular} \\
    \midrule
    Token MLP - S/16 \textsuperscript{1}     & 102 & 1.2 & 82.5 \\
    Channel MLP - S/16 \textsuperscript{2}  & 822 & 10.4 & 77.3 \\
    Token MLP - L/16 \textsuperscript{3}  & 411 & 7.5 & 55 \\
    \midrule
    2-Layer MLP \textsuperscript{4} & 1074 & 8.2 & 129.7 \\
    7-Layer MLP \textsuperscript{5} & 3.7 & 0.03 & 113.4 \\
    \bottomrule
  \end{tabular}

  \medskip

  \raggedright
  \footnotesize
  \textsuperscript{1}\,input: $[B\!\cdot\!C,\,T]=[512,196]$ with layer $196\!\rightarrow\!256\!\rightarrow\!196$. \\
  \textsuperscript{2}\,input: $[B\!\cdot\!T,\,C]=[196,512]$ with layer $512\!\rightarrow\!2048\!\rightarrow\!512$. \\
  \textsuperscript{3}\,input: $[B\!\cdot\!C,\,T]=[1024,196]$ with layer $196\!\rightarrow\!512\!\rightarrow\!196$. \\
  \textsuperscript{4}\,input $[256,1024]$ and hidden size $1024$ in all layers. \\
    \textsuperscript{5}\,input $[1,512]$ and hidden size $512$ in all layers. \\
\end{table}
\setlength{\tabcolsep}{6pt}

\begin{table*}[t!]
  \centering
  \caption{Comparison with prior AIE-based frameworks.}
  \label{tab:aie-full-comparison}
  \footnotesize
  \setlength{\tabcolsep}{5pt}
  \begin{tabular}{lcccccccc}
    \toprule
    \textbf{Framework}                   & \textbf{AIE Gen} & \textbf{Eff. (\%)} & \textbf{Fused Bias/Act} & \textbf{Wts On-AIE} & \textbf{Act On-AIE} & \textbf{Multi-Layer} & \textbf{Auto Place}  & \textbf{Max AIEs Used} \\
    \midrule
    \textbf{AIE4ML}               & AIEML/AIEMLv2    & 82.2         & \cmark             & \cmark            & \cmark               & \cmark               & \cmark                       & 296/304 (97.4\%)            \\
    AutoMM~\cite{automm}                 & AIE              & 27.5         & \xmark             & \xmark            & \xmark               & \cmark*               & \xmark                   & 192/400 (48\%)                    \\
    MaxEVA~\cite{maxeva}                 & AIE              & 56--60         & \xmark             & \xmark            & \xmark               & \xmark               & \xmark    
    &  400/400 (100\%)  \\
    GAMA~\cite{gamma}                    & AIEML            & 85         & \xmark             & \xmark            & \xmark               & \xmark               & \xmark                          & 288/304 (94.7\%)                    \\
    CHARM~\cite{10.1145/3686163}         & AIE              & 31            & \xmark             & \xmark            & \xmark               & \cmark*              & \xmark                         & 192/400  (48\%)                 \\
    ARIES~\cite{10.1145/3706628.3708870} & AIE              & 45       & \xmark             & \xmark            & \xmark               & \cmark*               & \cmark†                         & 320/400 (80\%)                 \\
    \bottomrule
  \end{tabular}
  \vspace{0.3em}
  \raggedright
  \footnotesize
  \\
  *supports multi-layer execution mainly via PL-side orchestration and buffering. \\
  †supports optional adaptive placement within user-defined core groups.
\end{table*}

\subsection{Comparison with Prior AIE-Based Frameworks}

Table~\ref{tab:aie-full-comparison} compares \texttt{AIE4ML} to prior AIE-based frameworks across architectural features, automation, and workload efficiency. Direct comparison is challenging as different works target different AIE generations, operate at different clock frequencies, and report metrics under different assumptions. To enable a fair deployment-relevant assessment, we report INT8 efficiency as a percent of each device’s peak performance, derived from reported sustained TOPS when available. When applicable, we provide an efficiency range that reflects multiple workloads or configurations reported by each framework.
Under a GEMM workload at full array utilization, as also reported in Subsection~\ref{subsec:model-performance}, \texttt{AIE4ML} sustains 82.2\% of the theoretical INT8 peak performance for the target AIE-ML device. Reported efficiency in most prior work is typically measured on isolated compute kernels; in contrast, \texttt{AIE4ML} includes on-chip data reordering performed by memory tiles, which is required for end-to-end execution.

\begin{table}[t]
  \centering
  \caption{End-to-end INT8 inference throughput for the 7-layer MLP used for cross-device comparison.}
  \label{tab:mlp-throughput-compare}
  \begin{tabular}{lccc}
    \toprule
    \textbf{Device}                 & \textbf{Generation} & \textbf{Toolchain}       & \begin{tabular}{c}\textbf{Throughput} \\ (TOPS)\end{tabular} \\
    \midrule
    Versal VEK280                   & AIE-ML              & AIE4ML                   & 113.4 \\
    VU13P FPGA                      & UltraScale+         & hls4ml                   & 3.7  \\
    Nvidia 3060 GPU                    & Ampere              & TensorRT                 & 14.1 \\
    Apple M4 ANE                    & 2024                & Core~ML                  & 10.5 \\
    \bottomrule
  \end{tabular}
\end{table}

\subsection{Cross-Architecture Comparison}

For cross-platform benchmarking, we employ the same 7-layer $512 \times 512$ MLP evaluated earlier in Table~\ref{tab:mixer-layers}, which provides a balanced workload that all accelerator platforms can execute efficiently. To provide a fair basis for comparison across architectures with different execution models, we select (i) batch sizes large enough to expose steady-state compute behavior, (ii) avoid extremely small workloads that would underutilize any device, and (iii) utilize state-of-the-art libraries for model deployment on each device. All measurements are based on 8-bit weights and 8-bit activations to ensure a common precision level across devices. This MLP serves as a neutral workload for throughput-oriented measurements that allows us to compare device capabilities. On AIE and FPGA designs we measure the cycle-accurate throughput reported by the Vitis toolchain which closely corresponds to the delivered performance in deployments where data arrive from PL. GPU (NVIDIA RTX 3060) and Apple ANE (Apple Neural Engine) operate via vendor-managed inference runtimes; for these devices we report throughput with both inputs and weights resident on the device, matching common benchmarking practice for ML accelerators. Table~\ref{tab:mlp-throughput-compare} summarizes the steady-state throughput for each device. 
\texttt{AIE4ML} achieves the highest throughput among all evaluated accelerators, reaching 113.4~TOPS and outperforming state-of-the-art GPU, FPGA, and ANE implementations by large margins. Although the GPU, FPGA and ANE baselines possess lower theoretical INT8 peaks (roughly  $50\%$, $19\%$, and $19\%$ of AIE\textendash ML, respectively), \texttt{AIE4ML} converts architectural potential into realized performance more effectively, as evidenced by its sustained throughput relative to peak compute capacity when executing the evaluated workload.

\section{Conclusion}
In this work, we presented \texttt{AIE4ML}, an end-to-end framework for compiling complete neural networks into optimized firmware for AMD's AIE-ML architecture. The proposed design exploits key architectural features of AIE-ML devices to sustain high utilization across both single-tile and multi-tile deployments. \texttt{AIE4ML} advances prior work by producing bit-exact, device-aware implementations directly from high-level ML toolflows. Experimental results demonstrate scalable, high-throughput inference performance. Future work will extend \texttt{AIE4ML} to support additional operators and model classes, providing a foundation for further research on emerging AIE-ML and AIE-MLv2 architectures.

\section*{Acknowledgment}

 This work has been funded by the Eric \& Wendy Schmidt Fund for Strategic Innovation through the CERN Next Generation Triggers project under grant agreement number SIF-2023-004. M.K. is supported by the US
 Department of Energy (DOE) under Grant No. DE-AC02-76SF00515. S.D. acknowledges funding from the German Federal Ministry for Research, Technology and Space (BMFTR).

\bibliographystyle{IEEEtran}
\bibliography{my_bib}

\end{document}